\documentclass[conference]{IEEEtran}
\IEEEoverridecommandlockouts

\usepackage{cite}
\usepackage{amsmath,amssymb,amsfonts}
\usepackage{algorithmic}
\usepackage{graphicx}
\usepackage{textcomp}
\usepackage{xcolor}
\usepackage{booktabs}
\usepackage{multirow}
\usepackage{hyperref}
\providecommand{\IEEEauthormark}[1]{\textsuperscript{#1}}
\def\BibTeX{{\rm B\kern-.05em{\sc i\kern-.025em b}\kern-.08em
    T\kern-.1667em\lower.7ex\hbox{E}\kern-.125emX}}
    
\begin{document}

\title{RF-HiT: Rectified Flow Hierarchical Transformer for General Medical Image Segmentation
}

\author{
    \IEEEauthorblockN{Ahmed Marouane Djouamaa\IEEEauthormark{$^1$}, Abir Belaala\IEEEauthormark{$^1$}, Abdellah Zakaria Sellam\IEEEauthormark{$^2$}, Salah Eddine Bekhouche\IEEEauthormark{$^3$},\\ Cosimo Distante\IEEEauthormark{$^2$}, and Abdenour Hadid \IEEEauthormark{$^4$}}\\
    
    \IEEEauthorblockA{\IEEEauthormark{$^1$} Dept. of Computer Science, Mohamed Khider University, Biskra, Algeria}
    \IEEEauthorblockA{\IEEEauthormark{$^2$} Inst. of Applied Sciences and Intelligent Systems, National Research Council of Italy, 73100 Lecce, Italy}
    \IEEEauthorblockA{\IEEEauthormark{$^3$} Dept. of Computer Science and AI, University of the Basque Country, UPV/EHU, San Sebastian, Spain}
    \IEEEauthorblockA{\IEEEauthormark{$^4$} Faculty of Data Science and Computing, Universiti Malaysia Kelantan, Kelantan, Malaysia}
    
}

\maketitle

\begin{abstract}
Accurate medical image segmentation requires both long-range contextual reasoning and precise boundary delineation, a task where existing transformer- and diffusion-based paradigms are frequently bottlenecked by quadratic computational complexity and prohibitive inference latency. We propose RF-HiT, a Rectified Flow Hierarchical Transformer that integrates an Hourglass Transformer backbone with a multi-scale hierarchical encoder for anatomically guided feature conditioning. Unlike prior diffusion-based approaches that rely on hundreds of denoising steps, RF-HiT leverages rectified flow with efficient transformer blocks, achieving linear complexity and requiring only a few discretization steps. The model further fuses conditioning features at each resolution via learnable interpolation, enabling effective multi-scale feature integration with minimal computational overhead. As a result, RF-HiT achieves a strong efficiency--performance trade-off, requiring only 10.14 GFLOPs, 13.6M parameters, and inference in as few as 3 steps. Despite its compact design, RF-HiT attains 91.27\% mean Dice on ACDC and 87.40\% on BraTS 2021, achieving performance comparable to or exceeding that of significantly more intensive architectures. These results suggest that RF-HiT is a promising, computationally efficient foundation for clinical image segmentation.
\end{abstract}


\section{Introduction}
Accurate medical image segmentation is fundamental to clinical decision-making and disease monitoring. This task demands pixel-level precision across highly heterogeneous imaging conditions. Manual annotation by clinical experts is time-consuming, subject to inter-rater variability, and does not scale to modern clinical workloads. These challenges have driven sustained interest in automated deep learning-based segmentation.

Convolutional neural networks (CNNs), particularly the U-Net architecture~\cite{ronneberger2015u}, have long dominated medical image segmentation. While U-Net effectively fuses local and contextual information, the inductive biases of convolution limit the model's receptive field, hindering the modeling of long-range spatial dependencies.

The Vision Transformer (ViT)~\cite{dosovitskiy2020image} demonstrated that the Transformer architecture~\cite{vaswani2017attention}, originally designed for sequence modeling, can effectively capture long-range dependencies in computer vision. By enabling global self-attention, ViT overcomes the limited receptive field of convolution, motivating its adaptation to dense prediction tasks~\cite{chen2021transunet, hatamizadeh2022unetr, hatamizadeh2021swin}. However, standard transformer-based segmentation models face a fundamental trade-off: global attention scales quadratically and is computationally prohibitive for high-resolution inputs, while local attention (e.g., Swin Transformer~\cite{liu2021swin}) restricts long-range communication.

To address this trade-off, the recently introduced Hourglass Diffusion Transformer (HDiT)~\cite{crowson2024scalable} effectively overcomes both the quadratic complexity of standard Transformers and the restricted receptive fields of CNNs. By utilizing global self-attention at a low-resolution bottleneck to capture global context, and employing neighborhood attention~\cite{hassani2023neighborhood} at higher-resolution scales to extract fine-grained local features, HDiT achieves linear complexity while preserving global context. Building on these advantages, we adopt the HDiT backbone for medical image segmentation.

In parallel, generative paradigms have achieved state-of-the-art performance in modeling complex distributions while providing uncertainty estimates that are unavailable to discriminative models. Denoising diffusion probabilistic models (DDPMs)~\cite{ho2020denoising} formulate segmentation as a conditional generation problem~\cite{wolleb2022diffusion, wu2024medsegdiff}, producing multiple plausible segmentation hypotheses and uncertainty maps through stochastic sampling. However, DDPMs require hundreds of inference steps, making them impractically slow for clinical deployment. Rectified flow~\cite{liu2022flow} offers a principled alternative, learning a neural ordinary differential equation (ODE) whose velocity field transports noise to data along straight-line trajectories, achieving high generation quality in as few as 10--20 steps.

Motivated by the inference efficiency of rectified flow and the computational advantages of HDiT, we propose the \textbf{R}ectified \textbf{F}low \textbf{Hi}erarchical \textbf{T}ransformer (\textbf{RF-HiT}), a novel and efficient generative framework for 2D medical image segmentation. RF-HiT features a dual-pathway architecture comprising a main flow network and a feature encoder. The main flow network employs a hierarchical encoder-decoder structure to predict the velocity field of the generative flow, applying neighborhood attention at high-resolution levels while restricting quadratic global self-attention to the lowest-resolution bottleneck. Meanwhile, the feature encoder mirrors the encoder path of the main network to extract multi-scale anatomical representations from the original image. Fusing these features into corresponding levels of the flow network injects structural context into the generation process, improving boundary localization while reducing transformer attention bottlenecks.

The resulting architecture is remarkably lightweight, comprising only 13.6 million parameters and requiring just 10.14 GFLOPs. Despite this compact footprint, RF-HiT achieves competitive performance against state-of-the-art deterministic and generative methods on two challenging benchmarks: the BraTS 2021 multi-region brain tumor segmentation challenge and the ACDC cardiac segmentation dataset.

Our main contributions are as follows:
\begin{itemize}
    \item We propose \textbf{RF-HiT}, a novel generative framework for medical image segmentation that integrates an $\mathcal{O}(n)$ hierarchical Hourglass Transformer backbone with rectified flow. We introduce a dual-pathway design featuring a main flow network paired with a hierarchical feature encoder, which injects multi-scale anatomical priors into the velocity field prediction.
    \item We achieve exceptional computational efficiency; our model requires only \textbf{13.6 million parameters} and \textbf{10.14 GFLOPs}, producing high-quality segmentation masks in just \textbf{3 sampling steps}.
    \item We validate RF-HiT on two diverse and challenging medical segmentation benchmarks: ACDC and BraTS 2021. Our compact model demonstrates performance comparable to or exceeding substantially heavier state-of-the-art methods.
\end{itemize}

\begin{figure*}[t]
\centering
\includegraphics[width=1.0\textwidth]{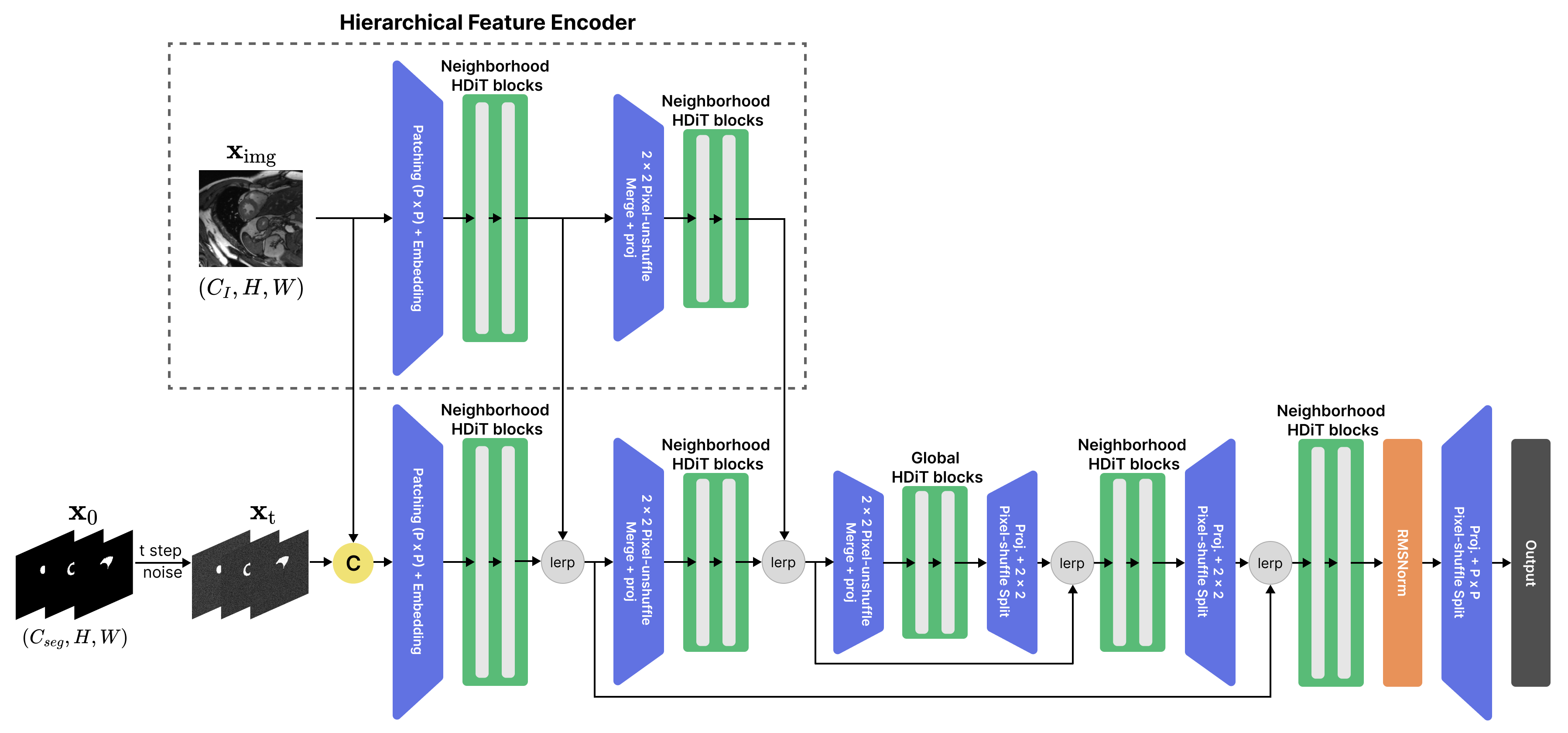}
\caption{Overview of the proposed RF-HiT: a rectified-flow segmentation framework composed of a main flow network that follows an encoder-decoder structure, where neighborhood attention is applied at high-resolution levels, and global attention at the bottleneck. A Hierarchical Feature Encoder provides multi-scale features that are fused into corresponding main network levels via learnable linear interpolation (lerp) to guide velocity prediction.}
\label{fig:architecture}
\end{figure*}

\section{Related Work}
\subsection{Generative Models for Medical Segmentation}
\label{sec:related_generative}
Continuous-time generative models, namely diffusion and flow-based approaches~\cite{ho2020denoising, liu2022flow}, have recently achieved state-of-the-art performance in image generation~\cite{peebles2023scalable}. They surpass GANs in image quality and diversity while avoiding their instability and mode collapse. Diffusion models corrupt data into Gaussian noise through a stochastic forward process, then train a neural network to approximate the reverse denoising process~\cite{ho2020denoising}. Amit et al.~\cite{amit2021segdiff} first applied diffusion models to image segmentation, proposing a dual-encoder architecture that merges features from the input image and current segmentation estimate before iteratively refining the segmentation through the denoising process. In medical imaging, Wolleb et al.~\cite{wolleb2022diffusion} applied diffusion models to brain tumor segmentation, conditioning the reverse process on MRI images as anatomical priors. Their stochastic sampling generates multiple segmentation hypotheses that, when averaged, improve segmentation performance while providing pixel-wise uncertainty maps. Subsequent work extended diffusion-based segmentation to volumetric data. Diff-UNet~\cite{xing2023diff} proposed the first end-to-end 3D diffusion model for medical segmentation, introducing a separate feature encoder for anatomical context and a Step-Uncertainty Fusion module that weights predictions across denoising steps. MedSegDiff-V2~\cite{wu2024medsegdiff} integrated vision transformers through Anchor Conditioning and a Spectrum-Space Transformer, achieving state-of-the-art performance across 20 medical segmentation tasks spanning five imaging modalities. Despite their effectiveness, diffusion models suffer from slow inference due to numerous denoising steps. Rectified flow~\cite{liu2022flow} offers a simpler alternative by learning a neural ODE that transports a source distribution to a target distribution along straight-line paths. These trajectories enable efficient generation with fewer steps. Recently, Bekhouche et al.~\cite{bekhouche2025segdt} incorporated rectified flow into a Diffusion Transformer for skin lesion segmentation, achieving strong performance in only 15 steps. However, rectified flow remains largely underexplored for broader medical segmentation.

\subsection{Medical Segmentation with Transformers}
Originally proposed for machine translation, the Transformer architecture~\cite{vaswani2017attention} has been widely adopted for computer vision~\cite{khan2022transformers}. Dosovitskiy et al.~\cite{dosovitskiy2020image} introduced the Vision Transformer (ViT), achieving competitive image recognition performance by treating images as sequences of patches. For medical image segmentation, Chen et al.~\cite{chen2021transunet} proposed TransUNet, a hybrid CNN-Transformer architecture combining CNN-based local feature extraction with Transformer-based global context modeling. To better exploit 3D volumetric data for brain tumor segmentation, Wang et al.~\cite{wang2021transbts} proposed TransBTS, which uses a 3D CNN to extract local features before modeling global dependencies across slices with a Transformer. Hatamizadeh et al.~\cite{hatamizadeh2022unetr} proposed UNETR, employing a pure ViT encoder that directly processes 3D patches and connects to a CNN decoder via skip connections to fuse local and global information. Standard ViT-based methods incur high computational cost for dense prediction due to the quadratic complexity of self-attention. To address this limitation, Liu et al.~\cite{liu2021swin} proposed the Swin Transformer, which uses shifted windows and hierarchical feature maps to efficiently capture multi-scale representations. Building on this, Hatamizadeh et al.~\cite{hatamizadeh2021swin} proposed Swin UNETR, which adopts a Swin Transformer encoder to extract multi-scale features and model long-range dependencies, followed by a CNN decoder for segmentation. More recently, the Hourglass Diffusion Transformer (HDiT)~\cite{crowson2024scalable} achieved remarkable success in high-resolution image synthesis. HDiT employs an efficient hierarchical architecture that applies neighborhood attention at higher-resolution stages, which Crowson et al.~\cite{crowson2024scalable} showed outperforms the shifted-window attention, and further incorporates learnable skip interpolation and 2D axial rotary positional embeddings to improve generation quality. Inspired by its success, we adapt the HDiT backbone as the basic unit of an efficient generative model for medical image segmentation.

\section{Method}
\subsection{Rectified Flow Models}
Rectified Flow~\cite{liu2022flow} learns an ODE transporting samples from a prior distribution $\pi_0$ to a target data distribution $\pi_1$. The key idea is to learn straight-line paths connecting the two distributions, yielding computationally efficient models that can be simulated accurately with few discretization steps.

Given a sample $\mathbf{X}_1$ from the data distribution and a noise sample $\mathbf{X}_0 \sim \mathcal{N}(0, \mathbf{I})$, we construct an interpolated sample $\mathbf{X}_t$ at time $t \in [0, 1]$ via linear interpolation:
\begin{equation}
    \mathbf{X}_t = t \mathbf{X}_1 + (1 - t) \mathbf{X}_0.
    \label{eq:interpolation}
\end{equation}
The ground-truth velocity, representing the direction of the straight path from noise to data, is:
\begin{equation}
    \mathbf{v}_t = \frac{\mathrm{d}\mathbf{X}_t}{\mathrm{d}t} = \mathbf{X}_1 - \mathbf{X}_0.
    \label{eq:velocity}
\end{equation}

A neural network $\mathbf{v}_\theta$ is trained to predict this velocity by minimizing the mean squared error:
\begin{equation}
    \mathcal{L} = \mathbb{E}_{t, \mathbf{X}_0, \mathbf{X}_1} \left\| \mathbf{v}_\theta(\mathbf{X}_t, t) - (\mathbf{X}_1 - \mathbf{X}_0) \right\|^2,
    \label{eq:rf_loss}
\end{equation}
where $t \sim \mathcal{U}(0, 1)$ is sampled uniformly. This objective is a simple least squares regression that can be minimized with standard gradient-based optimization.

At inference, we sample $\mathbf{X}_0 \sim \mathcal{N}(0, \mathbf{I})$ and solve the learned ODE forward in time. In practice, we use a first-order Euler solver with $N$ discrete steps. The straight-line formulation of Rectified Flow enables accurate sampling with relatively few steps compared to diffusion-based approaches.

\subsection{Efficient Hourglass Transformer Backbone}
    
Our velocity network $\mathbf{v}_\theta$ adapts the efficient Hourglass Diffusion Transformer (HDiT)~\cite{crowson2024scalable} from image synthesis to medical image segmentation. HDiT employs a hierarchical U-Net-like structure~\cite{ronneberger2015u} with an encoder-decoder structure connected via learnable skip connections. Representations are progressively downsampled via token-merging operations at each encoder level, processed at the lowest resolution in the bottleneck, and then upsampled symmetrically in the decoder. Skip connections between encoder and decoder levels at matching resolutions are merged using a learnable linear interpolation, enabling the model to adaptively balance fine-grained spatial details and high-level contextual features.
To enhance details and reduce computational cost, neighborhood self-attention is used at higher-resolution levels, while global self-attention is used at lower resolutions to enhance global coherence. This design achieves $\mathcal{O}(n)$ complexity with respect to the number of image tokens $n$, compared to the $\mathcal{O}(n^2)$ scaling of standard vision transformers.

Each transformer block incorporates AdaRMSNorm conditioning, where normalization scales are predicted by a mapping network from the timestep embedding.

\subsection{Proposed Framework}
As illustrated in Fig.~\ref{fig:architecture}, our overall framework consists of two parts: the main Hourglass Flow Network (HFN), and a separate Hierarchical Feature Encoder (HFE). HFN follows an encoder-decoder structure derived from HDiT. Specifically, we employ three hierarchical levels. The first two levels use neighborhood attention layers, while global self-attention is reserved for the final bottleneck level. HFE mirrors the encoder path of HFN. The two components operate as follows. First, HFN serves as our velocity network $\mathbf{v}_\theta$ and receives as input the channel-wise concatenation of the noised segmentation map $\mathbf{X}_t \in \mathbb{R}^{C_{\text{seg}} \times H \times W}$ and the original conditioning image $\mathbf{I} \in \mathbb{R}^{C_{\text{I}} \times H \times W}$, where $C_{\text{seg}}$ denotes the number of segmentation classes and $C_{\text{I}}$ the number of original image channels. $\mathbf{v}_\theta$ generates the predicted velocity field matching the spatial dimensions of $\mathbf{X}_t$. Second, HFE processes as input the original image $\mathbf{I}$ and extracts multi-scale features, which are fused with the corresponding encoder features of HFN at each hierarchical scale before downsampling. This fusion strengthens the image-guided segmentation refinement by introducing precise anatomical context from the conditioning image.
For feature fusion, we employ the same learnable linear interpolation mechanism originally used in HDiT for skip connections. Given main features $\mathbf{F}^{(l)}$ and conditioning image features $\mathbf{C}^{(l)}$ at level $l$, the fused representation is obtained through a learnable interpolation operation:
\begin{equation}
\tilde{\mathbf{F}}^{(l)} = (1-\alpha_l)\mathbf{F}^{(l)} + \alpha_l \mathbf{C}^{(l)},
\label{eq:lerp}
\end{equation}
where $\alpha_l$ is a trainable scalar parameter.
This per-level fusion injects rich, scale-consistent anatomical context into the generative process, providing stronger guidance for velocity field prediction and thereby improving boundary localization.

\section{Experimental Analysis}

\subsection{Datasets and Evaluation Metrics}
To evaluate the volumetric segmentation performance of our method, we utilize two publicly available segmentation datasets: ACDC and BraTS 2021. Moreover, the Dice Similarity Coefficient (DSC) and $95\%$ Hausdorff Distance (HD95) are adopted for quantitative comparison.

\textbf{ACDC.} The Automated Cardiac Diagnosis Challenge (ACDC) dataset~\cite{bernard2018deep} contains short-axis cine MR images from 150 patients. We use the 100 publicly available training cases, annotated for the left ventricle (LV), right ventricle (RV), and myocardium (Myo), and split them into 70, 10, and 20 cases for training, validation, and testing, respectively.

\textbf{BraTS 2021.} The BraTS 2021 dataset~\cite{baid2021rsna} contains 1,251 brain glioma cases. Each case includes four MRI modalities (T1, T1ce, T2, T2-FLAIR) resampled to 1\,mm$^3$ isotropic resolution with a volume size of $240\times240\times155$ voxels. Ground truth annotations denote three sub-regions, which are combined into three overlapping evaluation regions: Whole Tumor (WT), Tumor Core (TC), and Enhancing Tumor (ET). We split the dataset into training, validation, and test sets using a 70:10:20 ratio.

\subsection{Implementation Details}

All experiments are conducted using Python 3.10 and PyTorch 2.9.0 on a single NVIDIA TITAN V GPU. Table~\ref{tab:config} details the configurations of RF-HiT architecture. Input images are normalized and resized to $224 \times 224$. Furthermore, single-channel multi-class segmentation label maps are converted into multi-channel representations via one-hot encoding. Models are trained for 1000 epochs on ACDC and 300 epochs on BraTS 2021 with a batch size of 32. We use the AdamW optimizer with a learning rate of $1 \times 10^{-4}$ and weight decay of $1 \times 10^{-4}$. The learning rate follows a cosine annealing schedule with a linear warm-up phase spanning the first 1\% of training steps. To improve generalization, we apply a set of data augmentation techniques: random horizontal and vertical flips, random rotations and scaling, intensity scaling and shifting, gamma correction, and additive Gaussian noise. All 3D volumes are processed slice by slice, and the predicted 2D slices are stacked to reconstruct the 3D prediction for evaluation. Final segmentation masks are obtained by applying per-class probability thresholds selected via a grid search over 100 evenly spaced values in the interval $[0.2,0.8]$ on the validation set. We use $N=3$ Euler steps throughout, balancing segmentation quality and inference speed.

\begin{table}[t]
\centering
\caption{RF-HiT configurations.}
\label{tab:config}
\small
\begin{tabular}{lc}
\toprule
Hyperparameter & Value \\
\midrule
Patch Size & $4 \times 4$\\
Depth & $[2,2,2]$\\
Widths & $[128,256,384]$\\
Neighborhood Kernel Sizes & $[9, 13]$\\
Mapping Depth & 1 \\
Mapping Width & 256 \\
Mapping Hidden Dim & 784 \\
\bottomrule
\end{tabular}
\end{table}

\subsection{Comparison with State-of-the-Art Methods}
\subsubsection{ACDC Dataset}

Table~\ref{tab:acdc} reports segmentation performance on the ACDC cardiac MRI dataset for three classes: LV, Myo, and RV. Our approach achieves a highly competitive Average DSC of 91.27\%, ranking second overall just behind nnFormer (92.06\%), while utilizing significantly fewer parameters (13.6M vs 149.6M). Notably, our method achieves state-of-the-art segmentation performance on Myo (89.74\% DSC) and competitive results on RV (89.99\% DSC). The model's ability to simultaneously model local boundary cues via neighborhood attention and global shape context via bottleneck global attention appears particularly advantageous for this task, demonstrating that high efficiency does not compromise performance on complex cardiac structures.

\begin{table}[!h]
\centering
\caption{Comparison with state-of-the-art methods on ACDC dataset. \textbf{Bold}: best; \underline{underline}: second best.}
\label{tab:acdc}
\footnotesize
\setlength{\tabcolsep}{2pt}
\begin{tabular}{l l c c c c c c}
\toprule
\multirow{2}{*}{Type} & \multirow{2}{*}{Method} & \multirow{2}{*}{Params (M) $\downarrow$} & \multirow{2}{*}{GFLOPs $\downarrow$} & \multicolumn{4}{c}{DSC (\%) $\uparrow$} \\
\cmidrule(lr){5-8}
 & & & & RV & Myo & LV & Avg.\\
\midrule
\multirow{2}{*}{2D}
& TransUNet~\cite{chen2021transunet} & 96.07 & 88.91 & 88.86 & 84.53 & \underline{95.73} & 89.71 \\
& SwinUnet~\cite{cao2022swin} & \underline{27.17} & \textbf{6.2} & 88.55 & 85.62 & \textbf{95.83} & 90.00 \\
\midrule
\multirow{3}{*}{3D}
& MISSFormer~\cite{huang2021missformer} & 42.46 & \underline{7.25} & 89.55 & 88.04 & 94.99 & 90.86 \\
& UNETR~\cite{hatamizadeh2022unetr} & 92.58 & 54.70 & 85.29 & 86.52 & 94.02 & 88.61 \\
& nnFormer~\cite{zhou2023nnformer} & 149.6 & - & \textbf{90.94} & \underline{89.58} & 95.65 & \textbf{92.06} \\
\midrule
2D & \textbf{Ours} & \textbf{13.6} & 10.14 & \underline{89.99} & \textbf{89.74} & 94.09 & \underline{91.27} \\
\bottomrule
\end{tabular}
\end{table}

\subsubsection{BraTS 2021 Dataset}

Table~\ref{tab:brats21} presents a quantitative comparison of our method against state-of-the-art approaches on the BraTS 2021 dataset. Methods are evaluated using DSC (\%) and HD95 (mm) on the three standard tumor sub-regions: ET, TC, and WT. Our method achieves a mean DSC of 87.40\% and a mean HD95 of 5.08 mm, demonstrating solid performance on par with several recent models. While some fully 3D architectures like Swin UNETR and highly parameterized diffusion pipelines like MedSegDiff-V2 attain superior absolute performance, they impose a severe computational burden, demanding massive parameter counts and FLOPs, with MedSegDiff-V2 requiring 983 GFLOPS. In contrast, RF-HiT maintains highly competitive segmentation quality across overlapping tumor regions, particularly on Whole Tumor (91.30\% DSC) and Tumor Core (87.07\% DSC), while requiring substantially lower computational cost. With just 13.6 million parameters, our method provides an exceptional tradeoff between model size, inference efficiency, and segmentation accuracy.

\begin{table}[t]
\centering
\caption{Comparison with state-of-the-art methods on BraTS 2021. \textbf{Bold}: best; \underline{underline}: second best.}
\label{tab:brats21}
\setlength{\tabcolsep}{2pt}
\resizebox{\columnwidth}{!}{%
\begin{tabular}{llcccccc|cccc}
\toprule
\multirow{2}{*}{Type} & \multirow{2}{*}{Method} & \multirow{2}{*}{Params (M) $\downarrow$} & \multirow{2}{*}{GFLOPs $\downarrow$} & \multicolumn{4}{c|}{DSC (\%) $\uparrow$} & \multicolumn{4}{c}{HD95 (mm) $\downarrow$} \\
\cmidrule(lr){5-8}\cmidrule(lr){9-12}
 & & & & ET & TC & WT & Avg. & ET & TC & WT & Avg.\\
\midrule
\multirow{2}{*}{2D}
& TransUNet~\cite{chen2021transunet}     & 96.07 & \underline{88.91} & - & - & - & 86.6 & - & - & - & 13.74\\
& MedSegDiff-V2~\cite{wu2024medsegdiff}  & 46 & 983 & - & - & - & \underline{90.8} & - & - & - & 7.53\\
\midrule
\multirow{4}{*}{3D}
& SwinUnet3D~\cite{cai2023swin}          & 33.7 & - & 83.4 & 86.6 & 90.5 & 86.83 & - & - & - & -\\
& DiffBTS~\cite{nie2025diffbts}          & \underline{27.0} & - & \underline{86.33} & 90.70 & \underline{92.95} & 89.99 & \textbf{2.08} & \textbf{1.67} & \textbf{2.03} & \textbf{1.93}\\
& Swin UNETR~\cite{hatamizadeh2021swin}  & 61.98 & 394.84 & \textbf{89.1} & \underline{91.7} & \textbf{93.3} & \textbf{91.3} & - & - & - & -\\
& SegTransVAE~\cite{pham2022segtransvae} & 44.7 & 607.5 & 85.48 & \textbf{92.60} & 90.52 & 89.53 & \underline{2.89} & 5.84 & \underline{3.57} & \underline{4.10}\\
\midrule
2D & \textbf{Ours}                        & \textbf{13.6} & \textbf{10.14} & 83.83 & 87.07 & 91.30 & 87.40 & 4.16 & \underline{4.93} & 6.16 & 5.08\\
\bottomrule
\end{tabular}%
}
\end{table}

\subsection{Ablation Study}
\subsubsection{Effect of Hierarchical Feature Encoder (HFE)}
To investigate the contribution of HFE, we perform an ablation study comparing full RF-HiT against a baseline omitting HFE. Without HFE, the conditioning image is concatenated with the noisy mask at the HFN input, omitting multi-scale feature fusion. As shown in Table~\ref{tab:ablation_hfe}, incorporating the HFE significantly improves the segmentation performance, increasing the mean DSC from 90.69\% to 91.27\%. This gain highlights the limitations of standard input-level conditioning, where the main flow network must expend representational capacity to independently extract and propagate structural cues throughout its depth. By extracting dedicated multi-scale representations and explicitly infusing them at matching resolutions in the flow network, the HFE provides critical structural guidance that ensures local details and complex tissue boundaries are faithfully preserved during the generative process.

\begin{table}[t]
\centering
\caption{Ablation study on the effect of the Hierarchical Feature Encoder (HFE) evaluated on the ACDC dataset.}
\label{tab:ablation_hfe}
\small
\begin{tabular}{lc}
\toprule
Model Variant & Avg. DSC (\%) $\uparrow$\\
\midrule
w/o HFE & 90.69\\
\textbf{w/ HFE (Ours)} & \textbf{91.27}\\
\bottomrule
\end{tabular}
\end{table}

\subsubsection{Effect of Feature Fusion Mechanism}
To evaluate the impact of how HFE's features are injected into HFN, we compare the learnable linear interpolation (lerp) mechanism against a simple feature addition. In the addition variant, multi-scale conditioning features are directly added to the corresponding HFN representations. As shown in Table~\ref{tab:ablation_fusion}, simple addition yields a mean DSC of 91.09\% on the ACDC dataset. Learnable interpolation improves this to 91.27\% by adaptively weighting HFE conditioning features against HFN's internal representations at each hierarchical level. This demonstrates that parameterized fusion better balances anatomical conditioning against the model's internal generative representations than naive addition.

\begin{table}[t]
\centering
\caption{Ablation study comparing feature fusion mechanisms between HFE and HFN evaluated on the ACDC dataset.}
\label{tab:ablation_fusion}
\small
\begin{tabular}{lc}
\toprule
Fusion Mechanism & Avg. DSC (\%) $\uparrow$\\
\midrule
addition& 91.09\\
Learnable Linear Interpolation \textbf{(Ours)}& \textbf{91.27}\\
\bottomrule
\end{tabular}
\end{table}

\subsection{Visualizations}
\begin{figure}[t]
\centering
\includegraphics[width=1.0\columnwidth]{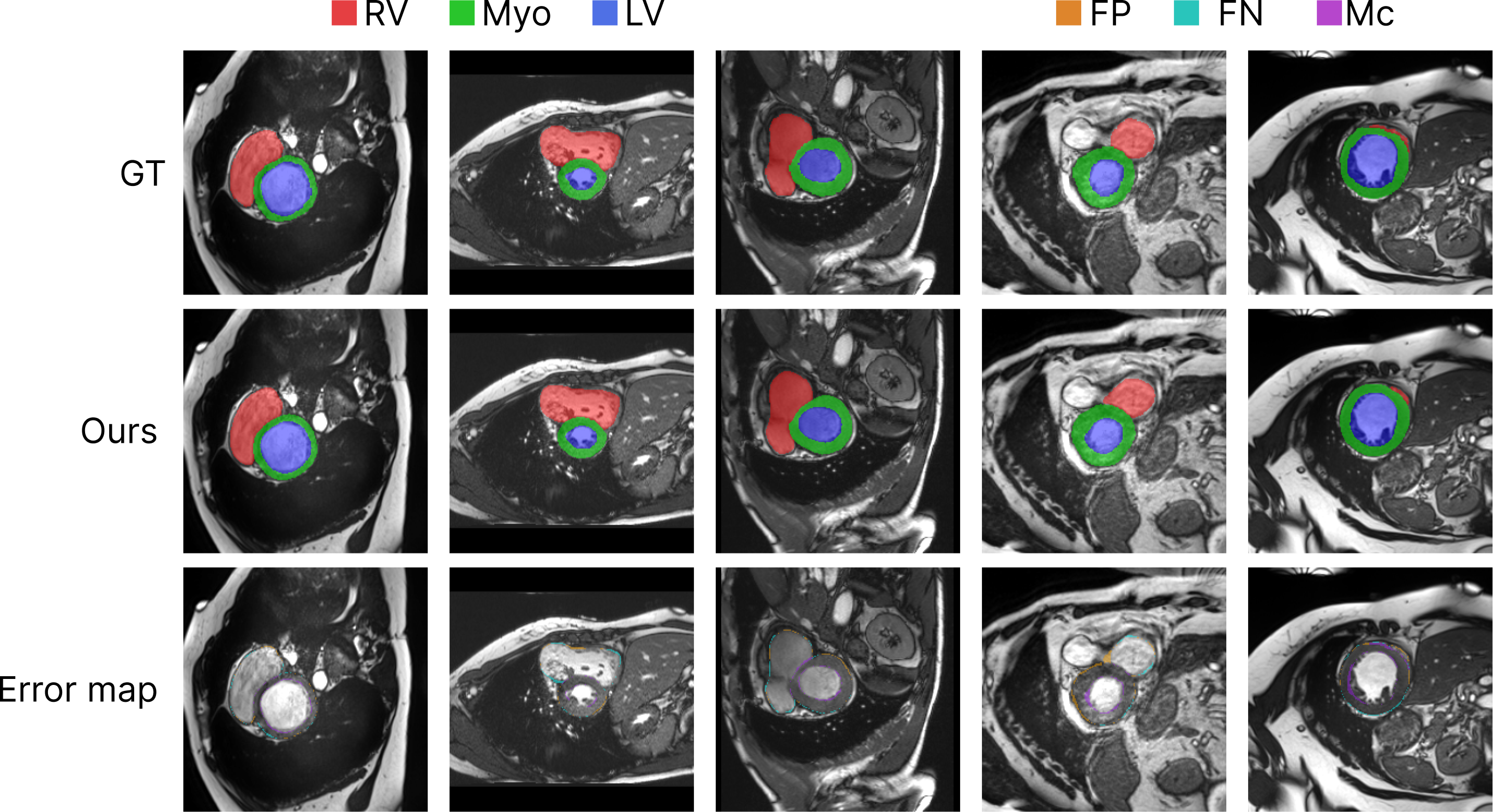}
\caption{Visualization of RF-HiT segmentation results on the ACDC dataset. Error maps highlight false positives (FP), false negatives (FN), and misclassifications (Mc).}
\label{fig:visualization}
\end{figure}

In Fig.~\ref{fig:visualization}, we present representative segmentation results by RF-HiT on the ACDC dataset. The final row displays pixel-wise error maps highlighting false positives, false negatives, and misclassification. The figure showcases highly accurate segmentations (first 3 columns) alongside more challenging anatomical cases (last 2 columns). Overall, RF-HiT consistently produces anatomically coherent segmentations with smooth boundaries. The predicted label maps exhibit strong spatial contiguity, forming continuous, well-connected regions for each anatomical class without spurious fragmentation. The error maps reveal that minor misclassifications occur primarily along anatomical boundaries, most notably between the Myo and LV. In cases where the RV is relatively small or geometrically complex (fifth column), confusion can also arise between the RV and Myo. Furthermore, in challenging scenarios exhibiting low tissue contrast or ambiguous structural boundaries (fourth column), the model occasionally struggles to delineate precise edges, leading to slight over-segmentation of the RV. Despite these edge-case difficulties, RF-HiT demonstrates robust generalization across a wide spectrum of cardiac morphologies.

\section{Conclusion}
In this work, we introduced RF-HiT, a Rectified Flow Hierarchical Transformer for efficient and accurate medical image segmentation. By combining an Hourglass Transformer backbone with multi-scale hierarchical feature conditioning, the proposed model effectively captures both global context and fine-grained structural details. The use of rectified flow enables fast inference with only a few discretization steps, while efficient transformer blocks ensure linear computational complexity. Extensive experiments on ACDC and BraTS 2021 demonstrate that RF-HiT achieves a strong balance between performance and efficiency, reaching competitive segmentation accuracy with significantly reduced computational cost. These results highlight the potential of rectified flow as a computationally practical alternative to diffusion models in medical imaging, paving the way for efficient generative segmentation in clinical settings.

\bibliographystyle{IEEEtran}
\bibliography{references}

@inproceedings{ronneberger2015u,
  title={U-net: Convolutional networks for biomedical image segmentation},
  author={Ronneberger, Olaf and Fischer, Philipp and Brox, Thomas},
  booktitle={International Conference on Medical image computing and computer-assisted intervention},
  pages={234--241},
  year={2015},
  organization={Springer}
}

@article{vaswani2017attention,
  title={Attention is all you need},
  author={Vaswani, Ashish and Shazeer, Noam and Parmar, Niki and Uszkoreit, Jakob and Jones, Llion and Gomez, Aidan N and Kaiser, {\L}ukasz and Polosukhin, Illia},
  journal={Advances in neural information processing systems},
  volume={30},
  year={2017}
}

@article{bernard2018deep,
  title={Deep learning techniques for automatic MRI cardiac multi-structures segmentation and diagnosis: is the problem solved?},
  author={Bernard, Olivier and Lalande, Alain and Zotti, Clement and Cervenansky, Frederick and Yang, Xin and Heng, Pheng-Ann and Cetin, Irem and Lekadir, Karim and Camara, Oscar and Ballester, Miguel Angel Gonzalez and others},
  journal={IEEE transactions on medical imaging},
  volume={37},
  number={11},
  pages={2514--2525},
  year={2018},
  publisher={ieee}
}

@article{dosovitskiy2020image,
  title={An image is worth 16x16 words: Transformers for image recognition at scale},
  author={Dosovitskiy, Alexey and Beyer, Lucas and Kolesnikov, Alexander and Weissenborn, Dirk and Zhai, Xiaohua and Unterthiner, Thomas and Dehghani, Mostafa and Minderer, Matthias and Heigold, Georg and Gelly, Sylvain and others},
  journal={arXiv preprint arXiv:2010.11929},
  year={2020}
}

@article{ho2020denoising,
  title={Denoising diffusion probabilistic models},
  author={Ho, Jonathan and Jain, Ajay and Abbeel, Pieter},
  journal={Advances in neural information processing systems},
  volume={33},
  pages={6840--6851},
  year={2020}
}

@article{chen2021transunet,
  title={Transunet: Transformers make strong encoders for medical image segmentation},
  author={Chen, Jieneng and Lu, Yongyi and Yu, Qihang and Luo, Xiangde and Adeli, Ehsan and Wang, Yan and Lu, Le and Yuille, Alan L and Zhou, Yuyin},
  journal={arXiv preprint arXiv:2102.04306},
  year={2021}
}

@inproceedings{wang2021transbts,
  title={Transbts: Multimodal brain tumor segmentation using transformer},
  author={Wang, Wenxuan and Chen, Chen and Ding, Meng and Yu, Hong and Zha, Sen and Li, Jiangyun},
  booktitle={International conference on medical image computing and computer-assisted intervention},
  pages={109--119},
  year={2021},
  organization={Springer}
}

@inproceedings{liu2021swin,
  title={Swin transformer: Hierarchical vision transformer using shifted windows},
  author={Liu, Ze and Lin, Yutong and Cao, Yue and Hu, Han and Wei, Yixuan and Zhang, Zheng and Lin, Stephen and Guo, Baining},
  booktitle={Proceedings of the IEEE/CVF international conference on computer vision},
  pages={10012--10022},
  year={2021}
}

@inproceedings{hatamizadeh2021swin,
  title={Swin unetr: Swin transformers for semantic segmentation of brain tumors in mri images},
  author={Hatamizadeh, Ali and Nath, Vishwesh and Tang, Yucheng and Yang, Dong and Roth, Holger R and Xu, Daguang},
  booktitle={International MICCAI brainlesion workshop},
  pages={272--284},
  year={2021},
  organization={Springer}
}

@article{amit2021segdiff,
  title={Segdiff: Image segmentation with diffusion probabilistic models},
  author={Amit, Tomer and Shaharbany, Tal and Nachmani, Eliya and Wolf, Lior},
  journal={arXiv preprint arXiv:2112.00390},
  year={2021}
}

@article{baid2021rsna,
  title={The rsna-asnr-miccai brats 2021 benchmark on brain tumor segmentation and radiogenomic classification},
  author={Baid, Ujjwal and Ghodasara, Satyam and Mohan, Suyash and Bilello, Michel and Calabrese, Evan and Colak, Errol and Farahani, Keyvan and Kalpathy-Cramer, Jayashree and Kitamura, Felipe C and Pati, Sarthak and others},
  journal={arXiv preprint arXiv:2107.02314},
  year={2021}
}

@article{huang2021missformer,
  title={Missformer: An effective medical image segmentation transformer},
  author={Huang, Xiaohong and Deng, Zhifang and Li, Dandan and Yuan, Xueguang},
  journal={arXiv preprint arXiv:2109.07162},
  year={2021}
}

@article{khan2022transformers,
  title={Transformers in vision: A survey},
  author={Khan, Salman and Naseer, Muzammal and Hayat, Munawar and Zamir, Syed Waqas and Khan, Fahad Shahbaz and Shah, Mubarak},
  journal={ACM computing surveys (CSUR)},
  volume={54},
  number={10s},
  pages={1--41},
  year={2022},
  publisher={ACM New York, NY}
}

@inproceedings{hatamizadeh2022unetr,
  title={Unetr: Transformers for 3d medical image segmentation},
  author={Hatamizadeh, Ali and Tang, Yucheng and Nath, Vishwesh and Yang, Dong and Myronenko, Andriy and Landman, Bennett and Roth, Holger R and Xu, Daguang},
  booktitle={Proceedings of the IEEE/CVF winter conference on applications of computer vision},
  pages={574--584},
  year={2022}
}

@article{liu2022flow,
  title={Flow straight and fast: Learning to generate and transfer data with rectified flow},
  author={Liu, Xingchao and Gong, Chengyue and Liu, Qiang},
  journal={arXiv preprint arXiv:2209.03003},
  year={2022}
}

@inproceedings{wolleb2022diffusion,
  title={Diffusion models for implicit image segmentation ensembles},
  author={Wolleb, Julia and Sandk{\"u}hler, Robin and Bieder, Florentin and Valmaggia, Philippe and Cattin, Philippe C},
  booktitle={International conference on medical imaging with deep learning},
  pages={1336--1348},
  year={2022},
  organization={PMLR}
}

@inproceedings{cao2022swin,
  title={Swin-unet: Unet-like pure transformer for medical image segmentation},
  author={Cao, Hu and Wang, Yueyue and Chen, Joy and Jiang, Dongsheng and Zhang, Xiaopeng and Tian, Qi and Wang, Manning},
  booktitle={European conference on computer vision},
  pages={205--218},
  year={2022},
  organization={Springer}
}

@inproceedings{pham2022segtransvae,
  title={Segtransvae: Hybrid cnn-transformer with regularization for medical image segmentation},
  author={Pham, Quan-Dung and Nguyen-Truong, Hai and Phuong, Nam Nguyen and Nguyen, Khoa NA and Nguyen, Chanh DT and Bui, Trung and Truong, Steven QH},
  booktitle={2022 IEEE 19th International Symposium on Biomedical Imaging (ISBI)},
  pages={1--5},
  year={2022},
  organization={IEEE}
}

@article{zhou2023nnformer,
  title={nnformer: Volumetric medical image segmentation via a 3d transformer},
  author={Zhou, Hong-Yu and Guo, Jiansen and Zhang, Yinghao and Han, Xiaoguang and Yu, Lequan and Wang, Liansheng and Yu, Yizhou},
  journal={IEEE transactions on image processing},
  volume={32},
  pages={4036--4045},
  year={2023},
  publisher={IEEE}
}

@inproceedings{peebles2023scalable,
  title={Scalable diffusion models with transformers},
  author={Peebles, William and Xie, Saining},
  booktitle={Proceedings of the IEEE/CVF international conference on computer vision},
  pages={4195--4205},
  year={2023}
}

@article{xing2023diff,
  title={Diff-unet: A diffusion embedded network for volumetric segmentation},
  author={Xing, Zhaohu and Wan, Liang and Fu, Huazhu and Yang, Guang and Zhu, Lei},
  journal={arXiv preprint arXiv:2303.10326},
  year={2023}
}

@inproceedings{hassani2023neighborhood,
  title={Neighborhood attention transformer},
  author={Hassani, Ali and Walton, Steven and Li, Jiachen and Li, Shen and Shi, Humphrey},
  booktitle={Proceedings of the IEEE/CVF conference on computer vision and pattern recognition},
  pages={6185--6194},
  year={2023}
}

@article{cai2023swin,
  title={Swin Unet3D: a three-dimensional medical image segmentation network combining vision transformer and convolution},
  author={Cai, Yimin and Long, Yuqing and Han, Zhenggong and Liu, Mingkun and Zheng, Yuchen and Yang, Wei and Chen, Liming},
  journal={BMC medical informatics and decision making},
  volume={23},
  number={1},
  pages={33},
  year={2023},
  publisher={Springer}
}

@inproceedings{crowson2024scalable,
  title={Scalable high-resolution pixel-space image synthesis with hourglass diffusion transformers},
  author={Crowson, Katherine and Baumann, Stefan Andreas and Birch, Alex and Abraham, Tanishq Mathew and Kaplan, Daniel Z and Shippole, Enrico},
  booktitle={Forty-first International Conference on Machine Learning},
  year={2024}
}

@inproceedings{wu2024medsegdiff,
  title={Medsegdiff-v2: Diffusion-based medical image segmentation with transformer},
  author={Wu, Junde and Ji, Wei and Fu, Huazhu and Xu, Min and Jin, Yueming and Xu, Yanwu},
  booktitle={Proceedings of the AAAI conference on artificial intelligence},
  volume={38},
  number={6},
  pages={6030--6038},
  year={2024}
}

@article{nie2025diffbts,
  title={Diffbts: A lightweight diffusion model for 3d multimodal brain tumor segmentation},
  author={Nie, Zuxin and Yang, Jiahong and Li, Chengxuan and Wang, Yaqin and Tang, Jun},
  journal={Sensors},
  volume={25},
  number={10},
  pages={2985},
  year={2025},
  publisher={MDPI}
}

@inproceedings{bekhouche2025segdt,
  title={SegDT: A Diffusion Transformer-Based Segmentation Model for Medical Imaging},
  author={Bekhouche, Salah Eddine and Maroun, Gaby and Dornaika, Fadi and Hadid, Abdenour},
  booktitle={International Conference on Image Analysis and Processing},
  pages={54--66},
  year={2025},
  organization={Springer}
}


\end{document}